\documentclass[preprint,12pt]{elsarticle}

\usepackage{amssymb}
\usepackage{amsmath}
\usepackage{times}
\usepackage{epsfig}
\usepackage{graphicx}

\usepackage{tikz}
\usepackage{float} 
\usepackage{algorithm}% http://ctan.org/pkg/algorithms
\usepackage{algpseudocode}% http://ctan.org/pkg/algorithmicx
\usepackage{booktabs}
\usepackage{amstext}
\usepackage{listings}

\usepackage[pagebackref,breaklinks,colorlinks]{hyperref}

\journal{Biomedical Informatics}

\begin{document}

\begin{frontmatter}



\title{Using Medical Algorithms for Task-Oriented Dialogue in LLM-Based Medical Interviews} %% Article title


\author[uminho]{Rui Reis}

\author[uminho]{Pedro Rangel Henriques}

\author[chsj]{João Ferreira-Coimbra}

\author[inesctec,ipca]{Eva Oliveira}

\author[inesctec,ipca]{Nuno F. Rodrigues\corref{cor1}}
\ead{nfr@ipca.pt}

\cortext[cor1]{Corresponding author}

\affiliation[uminho]{organization={University of Minho},
            city={Braga},
            country={Portugal}}

\affiliation[chsj]{organization={University Hospital Center of São João},
            city={Porto},
            country={Portugal}}

\affiliation[inesctec]{organization={INESC-TEC},
            city={Porto},
            country={Portugal}}

\affiliation[ipca]{organization={2Ai - School of Technology, IPCA},
            city={Barcelos},
            country={Portugal}}

%% Abstract
\begin{abstract}
\emph{Objective}: Hospital emergency departments face high patient volumes and time pressures, often compromising the collection of patient medical history. Traditional interviews risk incomplete or inconsistent data, impacting diagnostic accuracy. This study proposes a LLM-driven task-oriented dialogue system that uses medical algorithms and structured protocols to improve efficiency, adaptability, and quality of medical interviews.

\emph{Methods}: We developed a task-oriented dialogue framework structured as a Directed Acyclic Graph (DAG) of medical questions. The system integrates: (1) a systematic pipeline for transforming medical algorithms and guidelines into a clinical question corpus; (2) a cold-start mechanism based on hierarchical clustering to generate efficient initial questioning without prior patient information; (3) an expand-and-prune mechanism enabling adaptive branching and backtracking based on patient responses; (4) a termination logic to ensure interviews end once sufficient information is gathered; and (5) automated synthesis of doctor-friendly structured reports aligned with clinical workflows. Human-computer interaction principles guided the design of both the patient and physician applications. Preliminary evaluation involved five physicians using standardized instruments: NASA-TLX (cognitive workload), the System Usability Scale (SUS), and the Questionnaire for User Interface Satisfaction (QUIS).

\emph{Results}: The patient application achieved low workload scores (NASA-TLX = 15.6), high usability (SUS = 86), and strong satisfaction (QUIS = 8.1/9), with particularly high ratings for ease of learning and interface design. The physician application yielded moderate workload (NASA-TLX = 26) and excellent usability (SUS = 88.5), with satisfaction scores of 8.3/9. Both applications demonstrated effective integration into clinical workflows, reducing cognitive demand and supporting efficient report generation. Limitations included occasional system latency and a small, non-diverse evaluation sample.

\emph{Conclusion}: This study demonstrates the feasibility of using LLM-based task-oriented dialogue systems to streamline medical interviews. By combining medical algorithms with adaptive dialogue management, the system reduces clinician workload, improves data quality, and generates structured reports supportive of clinical decision-making. These findings suggest immediate benefits for emergency physicians working under high pressure and offer a foundation for developers of AI-driven healthcare systems to optimize patient interviews in diverse clinical contexts.
\end{abstract}



%% Keywords

\end{frontmatter}



\section{Introduction}\label{sec:Introduction__main}
Anamnesis, the process of gathering a patient’s medical history, is essential for accurate diagnosis and effective treatment. In high-pressure environments like emergency departments \cite{Silva2022,Silva2023}, this process is often compromised by systemic issues such as overcrowding, administrative burdens \cite{Cream2025Lost} and physician fatigue \cite{Johnson2016}. These challenges lead to errors, including incomplete data collection, lack of sufficient follow-up on patient responses, and premature conclusions.

Physicians are also susceptible to cognitive biases \cite{Johnson2016,Kunitomo2022}, especially when under stress or fatigue. These biases can result in overlooking or misinterpreting responses from patients who are elderly, intoxicated, or have communication challenges, as well as rushing to conclusions without thorough questioning. Such limitations reduce the quality of information collected, hindering accurate diagnosis and treatment planning.

The integration of Artificial Intelligence (AI), and specifically Large Language Models (LLMs), in healthcare workflows has opened new avenues for supporting and automating tasks like anamnesis. LLMs, such as ChatGPT, excel in generating human-like responses and have shown promise in structured, goal-oriented interactions like medical interviews \cite{Oliveira2023}. 

Over the past decades, a vast body of medical knowledge has been systematically gathered and curated in the form of diagnostic algorithms, clinical guidelines, and patient case reports. These resources represent a rich foundation for structuring the anamnesis process, yet their integration into automated systems remains limited. The central research question we pose in this work is how such curated knowledge can be leveraged to fine-tune and task-orient LLMs to conduct the entire medical interview process.

Simulating the structured question-and-answer format necessary to anamnesis presents several challenges that Large Language Models (LLMs) must address. Medical interviews are inherently exploratory, often requiring the system to deviate from a linear path. Patient responses can introduce unexpected directions, necessitating ``backtracking'' and exploring new branches of questioning. For example, a patient might initially report a headache, but subsequent answers could indicate abdominal symptoms, prompting a shift in focus. The system must maintain a task-oriented approach, adhering to the typical structure of anamnesis while remaining flexible and precise, avoiding unnecessary tangents.

Another critical challenge is the cold-start scenario, where no prior information about the patient is available. The system must define a broad yet efficient set of initial questions to gather foundational information, such as the patient’s chief complaint, medications, and allergies, while avoiding redundancy. It must also determine the optimal stopping point, ensuring enough critical information is collected without making the interaction unnecessarily lengthy. Finally, the generated medical reports must be concise, comprehensive, and doctor-friendly, summarizing the gathered data in a manner that supports clinical workflows and effective decision-making \cite{Ferreira2023,Pacheco2020}. These challenges demand a robust framework capable of balancing adaptability, efficiency, and relevance.

\paragraph{Contributions} This study explores the development of an AI-driven system powered by Large Language Models (LLMs) to enhance the efficiency and accuracy of data collection in emergency departments. The system employs a task-oriented dialogue framework with a graph-based structure, dynamically adapting to patient inputs to guide conversations through key medical topics. By demonstrating its effectiveness, this work contributes to the field of AI in healthcare, showcasing the potential of LLMs to support clinical workflows, reduce clinician workload, and improve patient care outcomes in high-demand environments.

The primary contributions of this work are as follows:

\begin{itemize}
\item We developed a systematic approach for processing medical algorithms, guidelines, and diagnostic protocols to construct a comprehensive corpus of clinically relevant questions.
\item We designed a novel cold-start mechanism that applies hierarchical clustering to medical questions, enabling effective initialization of interviews without prior patient information.
\item We introduced a dynamic expand-and-prune mechanism within the dialogue system, allowing conversations to flexibly adapt to patient responses while maintaining task-oriented flow.
\item We proposed a method to synthesize patient responses into structured, doctor-friendly medical reports, aligning with clinical workflows and supporting decision-making.
\end{itemize}

\section{Statement of significance}
\subsection{Problem or Issue} 
Collecting accurate and comprehensive patient histories (anamnesis) in emergency departments is time-consuming, error-prone, and impacted by overcrowding, administrative burdens, and physician fatigue.

\subsection{What is Already Known} 
Existing tools such as EMRs, static questionnaires, and conversational interfaces improve documentation and engagement but lack adaptability, often failing to capture nuanced, patient-specific information in real time.

\subsection{What this Paper Adds} 
This paper introduces a task-oriented dialogue framework powered by LLMs and structured as a dynamic graph of medical questions. It addresses cold-start challenges, adapts to patient responses through expand-and-prune logic, and generates concise, doctor-friendly reports. Preliminary evaluation with physicians shows improved efficiency, usability, and data quality.

\subsection{Who would benefit} Emergency physicians and developers of AI-driven healthcare systems seeking to optimize patient interviews and reduce clinician workload.

% =============

% =============

\section{Related Work}\label{sec:Related_Work__main}
The integration of informatics in anamnesis has driven significant advancements in healthcare. Electronic Medical Records (EMRs) enable structured documentation and streamlined data retrieval, enhancing the accuracy and accessibility of patient histories \citep{linoMedicalHistoryTaking2021}. While EMRs are effective for storage and retrieval, they do little to directly engage patients during the anamnesis process. Expanding on this, Internet-Based Anamnesis and Mobile Agent Applications allow patients to input medical information directly, increasing engagement and participation \citep{liuMobileAgentApplication2012,emmanouilAnamnesisInternetProspects2001}. However, these approaches often lack the adaptive, conversational structure necessary to capture nuanced, patient-specific information. This study addresses this gap by introducing a dynamic dialogue framework capable of tailoring interactions to patient responses in real time.

Conversational interfaces like “Talking to Ana” and digital self-assessment tools have further enhanced patient engagement, streamlining information collection and achieving high patient satisfaction \citep{deneckeTalkingAnaMobile2018,melmsPilotStudyPatient2021}. These tools often use empathy-driven questionnaires to explore the relationship between perceived empathy and patient outcomes \citep{martikainenPerceptionsDoctorsEmpathy2022}. While effective at fostering patient trust and satisfaction, these systems typically rely on static question sets, limiting their ability to adapt dynamically to new information provided by patients. By incorporating a task-oriented dialogue system, this approach enables flexible and context-sensitive questioning, mitigating the rigidity of static questionnaires.

Task-oriented dialogue systems have shown promise in managing structured, goal-oriented interactions across healthcare domains, such as mental health and chronic disease management. For example, systems in mental health use hierarchical Markov Decision Processes to respond empathetically to patient inputs \citep{papangelisAdaptiveDialogueSystem2013}, while chronic disease management systems employ deep reinforcement learning to adapt dynamically to patient responses in multi-turn conversations \citep{aliVirtualConversationalAgent2020}. However, these systems are often narrowly focused on specific conditions or domains, limiting their applicability in complex, multi-topic scenarios like emergency room anamnesis. This study addresses these limitations by designing a generalizable dialogue system capable of handling open-ended, multi-topic interactions.

While knowledge-graph-driven architectures have improved adaptability by enabling smooth transitions between topics, they remain constrained by high computational demands and limited scalability across diverse medical conditions \citep{sunKnowledgeGraphDriven2020}. Traditional slot-filling methods also struggle with unstructured inputs and complex branching dialogues. This work builds on these approaches by employing a graph-based dialogue framework that adapts dynamically to patient responses, ensuring comprehensive and contextually relevant information collection.

% =============

\section{System Design}\label{sec:System_Design__main}
We have developed a task-oriented dialogue (TOD) framework to structure medical interviews in a goal-driven manner for our AI-powered automated anamnesis system. Specifically designed for anamnesis, this framework ensures that patient interactions remain clinically relevant and diagnostic-focused. Our approach integrates a graph-based dialogue manager, Large Language Models (LLMs) for patient responses, a dynamic state tracking mechanism, and an adaptive decision-making module, allowing the system to guide conversations effectively and maintain a structured diagnostic flow.

\subsection{Graph Structure}\label{sec:Graph_Structure__main}
The task-oriented dialogue framework is structured as a Directed Acyclic Graph (DAG), denoted $G = (V, E)$ , organizing the dialogue flow in a logical, goal-oriented manner. This structure ensures the conversation progresses without cycles while avoiding redundant questions. The DAG is defined as follows:

\begin{itemize}
	\item \textbf{Nodes} $V$ : Each node $v \in V$ represents a specific medical question relevant to patient anamnesis, covering categories such as Symptoms, Medications, and Family History.
	\item \textbf{Edges} $E \subseteq V \times V$ : Directed edges connect pairs of nodes, with each edge $(v_i, v_j) \in E$ representing a logical transition from node $v_i$ to node $v_j$ based on the patient’s response. This structure ensures forward progression in the conversation, guiding it adaptively through medically relevant topics.
\end{itemize}

Each node $v \in V$ has specific attributes that define its role in the conversation. These attributes are stored on the nodes rather than the edges, as scenarios where $\text{rank}(v) > 1$ could make edge-level attributes more challenging to interpret.

\begin{itemize}
	\item \textbf{Priority} $P(v)$: The priority, selected from Low, Intermediate, High, or Urgent, dictates the order in which nodes are addressed. This priority is defined on node creation according to the patient's responses, ensuring urgent or contextually significant questions are prioritized.
	\item \textbf{State} $S(v)$: The state indicates the current status of the question within the conversation, with possible values of Open, Explore, and Closed. A node in the Open state has not yet been used to question the patient, while a node in the Explore state has been asked but may require follow-up questions. A node in the Closed state has been fully addressed and does not require further action.
	\item \textbf{Label} $L(v)$: The label denotes the medical area associated with each question. These labels, inspired by the work of \citep{wangUMASS_BioNLPMEDIQAChat20232023}, include the following: History of Present Illness, Review of Systems, Past Medical History, Medications, Chief Complaint, Past Surgical History, Allergy, Gynecologic History, Family History, Social History.
\end{itemize}

\subsubsection{Traversal Strategy}\label{sec:Traversal_Strategy__main}
Traversal through the Directed Acyclic Graph (DAG) is designed to guide the dialogue in a coherent and contextually relevant manner. The questions in the graph are organized into branches, with each branch representing a sequence of progressively specific questions about a particular health issue. As the traversal progresses down a branch (i.e., increases the distance from the initial vertex), the questions aim to confirm or clarify details related to the suspected health problem.

The system employs Depth-First Search (DFS) to maintain focus on a specific topic, such as cardiac symptoms, before shifting to another, like respiratory health. This approach minimizes context switching, reduces cognitive load for the patient, and ensures a logical flow of the conversation. By focusing in one topic at a time, DFS helps maintain continuity, keeping the patient engaged and improving the quality of the information gathered.

\begin{figure}[h]
	\centering
	\includegraphics[width=1.0\linewidth]{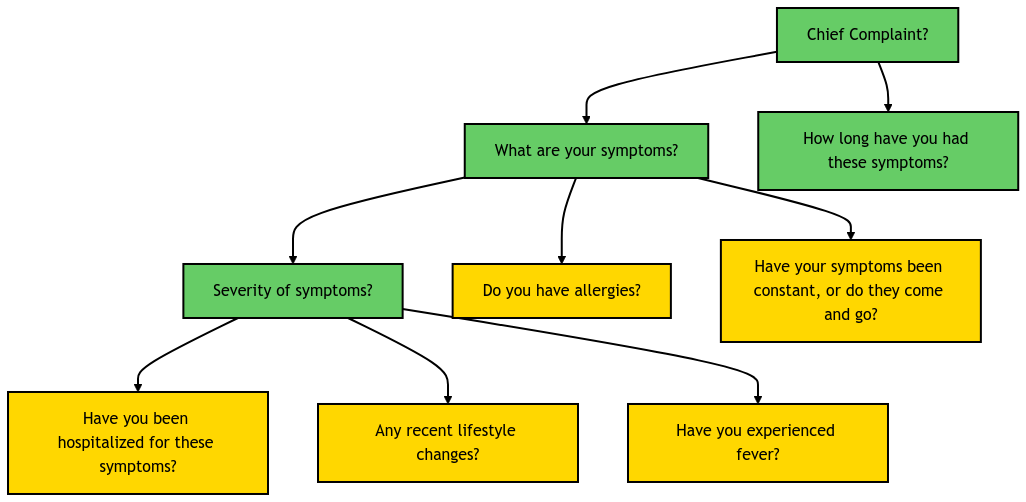}
	\caption{Example of a Direct Acyclic Graph (DAG) generated by the system, illustrating the sequence of questions and their relationships.}
	\label{fig:example-dag}
\end{figure}

\autoref{fig:example-dag} illustrates an example DAG, where nodes are color-coded based on their state. Green nodes represent questions that have been answered, either generating follow-up questions ($S(v) = \text{Explore}$) or being fully resolved ($S(v) = \text{Closed}$). Yellow nodes indicate unanswered questions ($S(v) = \text{Open}$).

While Breadth-First Search (BFS) is theoretically possible, we found it disrupts focus by frequently switching between unrelated topics. DFS is preferred for maintaining a more cohesive and focused interaction in medical dialogues.

\subsection{Cold Start}\label{sec:Cold_Start__main}
At the start of a patient interaction, the system operates without prior medical data, requiring a foundational set of general yet high-priority questions to guide the dialogue. These questions ensure critical information is gathered early and allow the system to adapt dynamically based on patient responses. The Directed Acyclic Graph (DAG) $G$ initializes with standalone nodes, each representing a single question prioritized for diagnostic relevance.

\subsubsection{Data Collection and Processing}\label{sec:Data_Collection_and_Processing__main}
The initial questions were derived from 1,020 diagnostic algorithms from the American Family Physician (AFP) repository, covering specialties like cardiology, dermatology, and neurology \citep{AmericanFamilyPhysician}. These algorithms were parsed into structured formats, extracting key diagnostic steps as questions. Using a language model, concise, patient-friendly questions were generated, such as:

\begin{itemize}
	\item \textbf{Cardiovascular}: “Do you experience chest pain?” or “Do you have a family history of heart disease?”
	\item \textbf{Infections}: “Do you have a fever?” or “Have you felt unusually fatigued?”
\end{itemize}

Each question was paired with a clinically validated justification, ensuring alignment with established medical guidelines. These justifications, while not patient-facing, provide transparency for developers and clinicians.

\subsubsection{Clustering Process}\label{sec:Clustering_Process__main}
Using OpenAI’s \texttt{text-embedding-ada-002}, each question was embedded into a high-dimensional vector space to capture semantic meaning. Hierarchical K-Means clustering was applied to organize questions into groups:

\begin{enumerate}
	\item \textbf{Broad Clusters}: Initial clustering grouped questions into general categories, such as pregnancy-related concerns or infection symptoms.
	\item \textbf{Refined Clusters}: Subgroups formed within each category, targeting specific conditions or symptoms, such as cardiovascular questions focused on chest pain or shortness of breath.
	\item \textbf{Filtering}: Rare or highly specific clusters were excluded to ensure initial questions remained broadly applicable, such as excluding genetic condition-specific inquiries.
\end{enumerate}

\subsubsection{Final Question Set}\label{sec:Final_Question_Set__main}
From the clustering process, the most general question from each category was selected to serve as the initial nodes in the DAG. These questions ensure coverage of the main diagnostic areas while remaining applicable to a broad patient population. These questions are:

\begin{itemize}
	\item \textbf{History of Present Illness}: "Can you describe the symptoms you are currently experiencing and when they started?" ($P(v) = \text{High}$) and "Have you had any recent illnesses or infections?" ($P(v) = \text{Intermediate}$)
	\item \textbf{Review of Systems}: "Have you experienced any recent weight loss?" ($P(v) = \text{High}$)
	\item \textbf{Past Medical History}: "Do you have any chronic illnesses or conditions that you have been diagnosed with in the past?" ($P(v) = \text{Intermediate}$)
	\item \textbf{Medications}: "Are you currently taking any medications?" ($P(v) = \text{High}$)
	\item \textbf{Chief Complaint}: "What is the primary health issue or symptom that prompted you to seek medical attention today?" ($P(v) = \text{Urgent}$)
	\item \textbf{Past Surgical History}: "Have you had any recent surgeries or medical procedures?" ($P(v) = \text{Intermediate}$)
	\item \textbf{Allergy}: "Do you have any known allergies to medications?" ($P(v) = \text{High}$)
	\item \textbf{Gynecologic History}: "Are you currently pregnant or planning to become pregnant?" ($P(v) = \text{Intermediate}$)
	\item \textbf{Family History}: "Is there a history of any significant medical conditions or diseases in your family?" ($P(v) = \text{Intermediate}$)
	\item \textbf{Social History}: "Do you consume alcohol or use recreational drugs?" ($P(v) = \text{Low}$)
\end{itemize}

Each question’s priority was manually determined to ensure the most critical and informative topics are addressed first, focusing on urgent concerns and essential diagnostic information. Questions related to the chief complaint and medications are assigned the highest priority, as they provide immediate insights necessary for informed medical decisions at the start of the interview. Lower-priority questions, such as those related to lifestyle or social history, are deferred until more pressing medical details have been collected.

\subsection{Decision-Making at Nodes}\label{sec:Decision_Making_at_Nodes__main}
In task-oriented medical dialogue systems, effective decision-making at each node of the Directed Acyclic Graph (DAG) is crucial to ensure a thorough yet efficient patient interview. This process determines whether a line of questioning should be pruned (i.e., no further questions on that topic are needed) or expanded with follow-up questions. 

At each node, the system evaluates the patient’s response based on its completeness and relevance to determine the next step. This process is dynamically managed by a LLM, which analyzes the response in real time using three key inputs: the current question, the patient’s feedback, and the accumulated conversation history. The LLM assigns a priority to the node based on the response’s clarity and relevance, deciding whether to expand the dialogue with follow-up questions, mark the node as resolved, or backtrack to explore an alternative path in the graph. This ensures that the conversation remains adaptive and focused on gathering the most pertinent medical information. Moreover, the specific Hard Prompt used by the LLM is illustrated in Listing~\texttt{\ref{lst:llm_prompt}}.

\begin{figure}[!ht]
\centering
\begin{lstlisting}[label={lst:llm_prompt}, caption={Prompt used for guiding the LLM during decision making.}]
You're tasked with relating the dialogue above with a specific question. The answer to the question
should be very complex and require a lot of information. If necessary, you can ask follow up questions.
You can make any of the following decisions:
- Prune: The dialogue is sufficient to answer the question, and the question is no longer relevant - should only happen if the question is answered in its fullest extent.
- Expand: The dialogue is not enough to answer the question, and follow up questions must be derived.
The decision should be made very thoughtfully, think step by step, and be as specific as possible.
**Very Important**: Make sure to ask very informative follow up questions, but Avoid making follow up questions that have already been answered in the dialogue.
When you do make follow up questions, make sure to include a few questions, which all should be very closely related to the question at hand and the dialogue at hand. These questions are meant to be used for self-anamnesis, thus should align to retrieve the most information possible about the chief complaint, in the most efficient way possible - but without asking too many questions.
The question to be considered is the following:
{self.question}
\end{lstlisting}
\end{figure}

Pruning occurs when the LLM determines that the patient’s response provides sufficient information to address the question fully, with no additional details necessary. The system then closes the current node, marking it as \textbf{Closed} and moving on to the next relevant node according to the DAG’s structure. Pruning is applied to avoid redundancy and keep the conversation focused, thus reducing the likelihood of overwhelming the patient with repetitive questions.

On the other hand, expansion takes place when the patient’s response is either incomplete, ambiguous, or introduces new concerns that merit further exploration. In such cases, the LLM generates tailored follow-up questions, which the system adds to the DAG as new nodes, updating the current node’s state to Explore. For example, if a patient mentions experiencing discomfort without specifying details, the system will follow up with additional inquiries to clarify factors such as duration, intensity, or location. Expansion ensures that essential information is gathered comprehensively, adapting to new, relevant topics that may arise in real-time.

The node decision-making process is powered by the LLM.evaluate function, which operates at each node to analyze patient responses. This function takes as input the current question, the patient’s response, and the conversation history, and outputs a decision in a structured JSON format. To ensure the JSON structure is well-formed and adheres to a predefined schema, OpenAI’s function-calling capabilities are employed. These functions enable controlled LLM generation that strictly follows the specified JSON schema, ensuring consistency and ease of management.

The JSON output includes the following elements:

\begin{itemize}
	\item \texttt{type}: Specifies whether the node should be \textbf{‘prune’} (indicating no follow-up questions are generated and the node is resolved) or \textbf{‘expand’} (indicating that follow-up questions are required).
	\item \texttt{follow\_up\_questions}: A sequence of follow-up questions, each represented as a vertex with an assigned priority, label and question text.
\end{itemize}

This structured output facilitates seamless integration into the recursive decision-making function. Once a node has been fully addressed, the function advances to the next unanswered node, ensuring only pertinent questions are presented. This approach is detailed in Algorithm~\autoref{alg:decision_making_repeated}, providing a robust and adaptive framework for task-oriented dialogue management.

\begin{algorithm}[h]
\caption{Node Decision-Making Algorithm with Repeated Question Handling}\label{alg:decision_making_repeated}
\begin{algorithmic}
\Require Current node $v \in V$, Graph $G = (V, E)$, Patient response $r$, Conversation history $\mathcal{H}$
\Ensure Updated DAG $G$ with state $S(v)$ modified, avoiding repeated questions
\State decision $\gets$ \text{LLM.evaluate}($v$, $r$, $\mathcal{H}$) 
\If{$\text{decision}.\text{type} = \text{prune}$}
    \State Set $S(v) \gets \text{closed}$ 
    \State $v_{next} \gets \text{succ}\{v \in V \mid S(v) = \text{open}\}$
    \If{$v_{next} = \text{null}$} 
        \State \textbf{return} $G$
    \EndIf
	\State Call \text{Node Decision-Making Algorithm} on $(v_{next}, G, r)$ 
\ElsIf{$\text{decision}.\text{type} = \text{expand}$}
    \State Set $S(v) \gets \text{explore}$ 
    \For{each question $q$ in decision.follow\_up\_questions}
        \If{$q \notin V$} 
            \State Add node $q$ to $V$
            \State Add edge $(v, q)$ to $E$
            \State Set $S(q) \gets \text{open}$
        \EndIf
    \EndFor
\EndIf
\State Output the updated DAG $G = (V, E)$
\end{algorithmic}
\end{algorithm}

This recursive algorithm allows the system to dynamically adapt its questioning strategy, continuously updating the DAG to reflect the conversation’s state.

\subsection{Termination Logic}\label{sec:Termination_Logic__main}
The termination logic ensures that the dialogue gathers comprehensive information without excessive length. The system uses a termination score to gauge when enough medical topics have been addressed. Each topic, represented by a unique node label in the DAG (e.g., medications, family history), is marked as covered once the system receives a sufficiently detailed response. The termination score is calculated by dividing the number of addressed topics by the total number of predefined topics:

\begin{equation}
	\text{Termination Score} = \frac{\text{\# Unique Node Labels Answered}}{\text{\# Unique Node Labels}}
\end{equation}

When this score meets a set threshold, typically 0.99 for thorough coverage, the conversation concludes. In routine cases, a lower threshold, such as 0.80, allows the dialogue to finish after covering the core topics.

To prevent the conversation from continuing indefinitely, a safeguard mechanism limits the total number of exchanges allowed. If this limit is reached before the termination score threshold, the conversation ends automatically, ensuring that the interaction remains efficient and avoids excessive questioning. This dual approach ensures that the dialogue collects necessary information while staying concise and focused.

\subsection{Medical Report Generation}\label{sec:Medical_Report_Generation__main}
The medical dialogue system compiles patient information into a structured medical report, providing healthcare professionals with a comprehensive yet concise summary of the patient’s health. Using the Directed Acyclic Graph (DAG) structure, each node in the DAG corresponds to a specific question, and the associated labels categorize these questions under distinct medical topics such as “Medications” or “Family History.” Throughout the conversation, the system tracks completed nodes, which contain fully provided responses, and organizes them by category. This approach mirrors the structure of professional medical notes, ensuring that all necessary information is captured and structured in a logical, organized manner.

The report generation process begins by grouping closed nodes by their labels, after which the system uses a Large Language Model (LLM) to generate bullet-point summaries for each category. For example, in the “Medications” category, the summary might list “Patient takes lisinopril for hypertension” and “No reported drug allergies.” Categories are prioritized based on the average importance of their nodes, with critical areas like Medications or Chief Complaint listed first. Additionally, the system generates a symptomatic summary - a brief, high-level overview of primary concerns and significant symptoms, such as “The patient reports controlled hypertension with no recent health changes or allergies."

The report generation process is formalized in Algorithm~\autoref{alg:generate_medical_report}, which details how the system organizes and prioritizes responses.

\begin{algorithm}
\caption{Medical Report Generation Using DAG}\label{alg:generate_medical_report}
\begin{algorithmic}
\Require Directed Acyclic Graph $G = (V, E)$, Patient gender
\Ensure Medical report containing lists of facts per category and a global symptomatic summary
\State $C \gets \{\}$ \Comment{Initialize dictionary to store questions and answers by category}
\For{each $v \in V$}
\If{$S(v) = \text{closed}$ or $S(v) = \text{explore}$}
\State $L \gets L(v)$ \Comment{Retrieve category label}
\If{$L \notin C$}
\State Initialize $C[L] \gets []$
\EndIf
\State Append $(v.\text{question}, v.\text{answer})$ to $C[L]$
\EndIf
\EndFor
\State $\text{MedicalReport} \gets \emptyset$ \Comment{Initialize medical report structure}
\For{each category $L \in \text{sorted keys of } C$}
\State $facts \gets \text{LLM.generateFacts}(C[L])$
\For{each fact in $facts$}
\State Append fact as a bullet point under $L$ in $\text{MedicalReport}$
\EndFor
\EndFor
\State $\text{Summary} \gets \text{LLM.generateSummary}(\{C[l] \mid l \in C \})$ \Comment{Summary using all questions and answers}
\State Append $\text{Summary}$ to $\text{MedicalReport}$
\State \Return $\text{MedicalReport}$ 
\end{algorithmic}
\end{algorithm}

This algorithm identifies relevant nodes (those in closed or explore states), organizes them by category, and uses an LLM to generate concise summaries, namely using the Hard Prompt in Listing~\texttt{\ref{lst:llm_report_prompt}}. Summaries are prioritized based on average priority $P(v)$ and the number of nodes within each category, ensuring that the most critical information appears first in the report.

\begin{figure}[!ht]
\centering
\begin{lstlisting}[label={lst:llm_report_prompt}, caption={Prompt used for generating medical reports with at LLM.}]
You're a master algorithm at summarizing key findings about a patient's health.
You're given a list of interactions between a nurse and a patient, which are composed by
a question and an answer, and you're asked to generate a the symptomatic summary of the patient's health.
The resulting summary should use medical terminology and be as concise as possible and should contain the most important information about the patient's health in a single small paragraph. The paragraph should be VERY short, no more than a two sentences.
The output language should be '{language}'.
The interactions are the following:
{nodes_repr}
\end{lstlisting}
\end{figure}

% =============

\section{Results}\label{sec:Results__main}
\subsection{Study Population}\label{sec:Study_Population__main}
The study involved five practicing physicians from Portugal. Of the participants, 20\% were male and 80\% were female. Participants’ ages ranged from 18 to 50, with 20\% in the 18–30 group and 80\% in the 30–50 group. On average, participants had 12 years of experience across various specialties and worked in hospitals and healthcare centers.

\subsection{Evaluation Instruments}\label{sec:Evaluation_Instruments__main}
The study utilized three evaluation tools:

\begin{itemize}
	\item NASA-TLX measured cognitive workload across dimensions such as Mental Demand, Effort, and Temporal Demand.
	\item The System Usability Scale (SUS) was used to assess usability through a series of ten standardized statements.
	\item QUIS evaluated satisfaction with interface elements such as Ease of Learning, Response Time, and Visual Layout.
\end{itemize}

\subsection{Results for the Patient Application}\label{sec:Results_for_the_Patient_Application__main}
\begin{itemize}
	\item \textbf{NASA-TLX}: The average workload score was 15.6. Temporal Demand had the lowest score, averaging 8, while Performance had the highest score of 22.
	\item \textbf{SUS}: The application scored an average of 86. Scores for ease of use (SUS03) were high, while frequent use desirability (SUS01) received lower scores.
	\item \textbf{QUIS}: The application scored 8.1 on average. Ease of Learning and Visual Layout scored 8.4, while Response Time scored 7.2.
\end{itemize}

\subsection{Results for the Physician Application}\label{sec:Results_for_the_Physician_Application__main}
\begin{itemize}
	\item \textbf{NASA-TLX}: The average workload score was 26. Effort scored the lowest at 18, while Mental Demand and Performance scored the highest at 34.
	\item \textbf{SUS}: The application scored an average of 88.5. SUS04, related to system simplicity, scored the lowest at 1.8.
	\item \textbf{QUIS}: The application scored 8.3 on average. Ease of Learning and General Functionality scored 8.6, while Response Time scored 8.0.
\end{itemize}

% =============

\section{Discussion}\label{sec:Discussion__main}
The results offer key insights into the performance of both the patient and physician applications. The patient application demonstrated low cognitive workload, with a NASA-TLX score of 15.6, and with a SUS score of 86. Participants rated Ease of Learning and Visual Layout highly on the QUIS (8.4 each), highlighting the system’s ease of use and intuitive design. However, Response Time (7.2) and Performance (22) pointed to occasional delays and a lack of user confidence in task outcomes. Despite these areas for improvement, the application demonstrated strong potential to streamline patient interactions and reduce cognitive demand.

The physician application indicated moderate cognitive workload, with a NASA-TLX score of 26. Higher scores for Mental Demand and Performance (34 each) suggest cognitive strain and lower confidence in task success. Usability was rated highly, with a SUS score of 88.5, though simplicity (SUS04: 1.8) indicated areas for workflow improvement. The QUIS score of 8.3 highlighted satisfaction with features like Ease of Learning and General Functionality (8.6 each), though Response Time (8.0) revealed occasional latency during tasks like report editing. The application effectively supported clinical workflows, enabling efficient review and editing of structured medical reports and integrating seamlessly with existing practices.

Overall, both applications demonstrated the potential to enhance usability and reduce manual workloads, enabling clinicians to focus more on patient care. Addressing areas such as system responsiveness and task confidence will further refine their effectiveness in clinical settings.

% =============

\section{Limitations and Future Work}\label{sec:Limitations_and_Future_Work__main}
This study, while promising, has several limitations that highlight opportunities for improvement and further research. The lack of large-scale, representative datasets for medical interviews was a critical limitation. This constraint hinders the validation and adaptability of the system across various medical domains. Future work should prioritize the development and curation of such datasets, which would not only enhance the system’s accuracy but also improve its reliability and applicability to specific clinical contexts.

One significant limitation was the cognitive load experienced in the physician application, along with suboptimal system responsiveness in both applications. These issues may have hindered usability and acceptance of the system. Future research should focus on refining the user interfaces and optimizing the underlying technology to ensure a smoother and more efficient experience.

Another limitation was the participant pool, which lacked representation from diverse user groups, particularly patients. This limitation affects the generalizability of the study’s findings. Expanding the participant pool in future studies to include more diverse demographics and user profiles would enable a more comprehensive evaluation of the applications’ effectiveness in real-world scenarios. Furthermore, future work could introduce configurable interaction modes to adapt the system for specific use cases, such as focusing exclusively on the chief complaint or conducting a comprehensive data collection. Additionally, implementing features that retain responses with long-term validity, such as family history, could reduce redundant questioning in follow-up interactions. Providing patients with a clear view of the process - such as displaying the key stages of the interview and indicating their progress - could also enhance user engagement and transparency.

Lastly, the current approach to determining when to end the interview, based on a global termination score, is another area that could be improved. Future work should explore topic-specific thresholds, allowing for more granular control of conversation depth across different medical categories. This could ensure thorough coverage of critical topics while avoiding unnecessary questions in less relevant areas. Additionally, investigating whether there is a practical upper limit to the number of questions asked - beyond which little new information is typically gathered - could further refine the system’s efficiency and prevent overburdening patients.

% =============

\section{Conclusion}\label{sec:Conclusion__main}
This study developed and evaluated an AI-driven system using Large Language Models (LLMs) to streamline medical interviews. The primary objectives were to improve the efficiency and accuracy of patient data collection, reduce clinician workload, and enhance medical reporting quality. By employing a task-oriented dialogue framework structured as a Directed Acyclic Graph (DAG), the system demonstrated its adaptability and relevance in clinical interactions.

The system successfully achieved its primary objectives, as shown by high usability and satisfaction scores. Despite limitations in responsiveness and participant diversity, the results highlight the potential of AI-powered tools to enhance clinical workflows and patient care in high-demand healthcare settings.

\section*{Ethics Statement}

All procedures involving human participants were conducted in accordance with 
the ethical standards of the institutional and national research committees, 
as well as with the 1964 Helsinki Declaration and its later amendments. 
The study protocol was reviewed and approved by the Ethical Committee of the 
University Hospital of São João (approval date: 2019; reference number: 393/19). 

Informed consent was obtained from all individual participants included in the study. 
The privacy rights of all human subjects were fully observed throughout the study.

\section*{Acknowledgments}
The authors acknowledge financial support from FCT - Fundação para a Ciência e Tecnologia, 
under the grant 2022.07391.PTDC.

\bibliographystyle{elsarticle-num} 
\bibliography{main}

\end{document}